\documentclass[english]{article}
\usepackage{eucal}
\usepackage[utf8]{inputenc}
\usepackage[T1]{fontenc}
\usepackage{babel}
\usepackage{amsmath}
\usepackage{wrapfig}
\usepackage[svgnames]{xcolor}
\definecolor{ocre}{RGB}{243,102,25}
\usepackage{caption}
\usepackage[font={color=ocre,bf},figurename=Fig.,labelfont={it}]{caption} 
\usepackage{float}
\usepackage{amsfonts}
\usepackage{csquotes}
\usepackage{amstext}
\usepackage{amssymb}
\usepackage{graphicx}
\usepackage{comment}
\usepackage{subcaption}
\usepackage{fancyhdr}
\usepackage{siunitx}
\usepackage{hyperref}
\hypersetup{
    colorlinks=true,
    linkcolor=blue,
    filecolor=magenta,      
    urlcolor=cyan,
}
\usepackage{graphicx}

\usepackage{biblatex}
\begin{document}


\title{Deep Neuroevolution Squeezes More out of Small Neural Networks and Small Training Sets: Sample Application to MRI Brain Sequence Classification}

\author{ \textbf{Joseph Stember}$^1$
\and 
\textbf{Hrithwik Shalu}$^2$}
\maketitle
\thispagestyle{fancy}
\noindent
\textsuperscript{1}Memorial Sloan Kettering Cancer Center, New York, NY, US, 10065 
\\
\textsuperscript{2}Indian Institute of Technology, Madras, Chennai, India, 600036
\\
\noindent
\textsuperscript{1}joestember@gmail.com
\\
\textsuperscript{2}lucasprimesaiyan@gmail.com 
\\


\begin{abstract}

\indent \textit{Purpose} Deep Neuroevolution (DNE) holds the promise of providing radiology artificial intelligence (AI) that performs well with small neural networks and small training sets. We seek to realize this potential via a proof-of-principle application to MRI brain sequence classification.

\indent \textit{Materials and Methods} We analyzed a training set of 20 patients, each with four sequences/weightings: T1, T1 post-contrast, T2, and T2-FLAIR. We trained the parameters of a relatively small convolutional neural network (CNN) as follows: First, we randomly mutated the CNN weights. We then measured the CNN training set accuracy, using the latter as the fitness evaluation metric. The fittest "child" CNNs were identified. We incorporated their mutations into the "parent" CNN. This selectively mutated parent became the next generation's parent CNN. We repeated this process for approximately 50,000 generations.

\indent \textit{Results} DNE achieved monotonic convergence to $100 \%$ training set accuracy. DNE also converged monotonically to 100 $\%$ testing set accuracy. 

\indent \textit{Conclusion} DNE can achieve perfect accuracy with small training sets and small CNNs. Particularly when combined with Deep Reinforcement Learning, DNE may provide a path forward in the quest to make radiology AI more human-like in its ability to learn. DNE may very well turn out to be a key component of the much-anticipated "meta-learning" regime of radiology AI; algorithms that can adapt to new tasks and new image types, similar to human radiologists. 
 
\end{abstract}

\pagebreak

\section*{Introduction}

\subsection*{Artificial intelligence in radiology currently focuses on specific tasks.}

Artificial intelligence (AI), and more specifically Deep Learning, exploded onto the scene of radiology research over the past decade. There is an exponentially increasing volume of publications in the field. AI in radiology typically makes use of convolutional neural networks (CNNs), a type of artificial neural network specifically suited for image feature detection. Despite the early promise and interest, currently, Radiology AI is focused on performing specific tasks, such as image classification, localization or segmentation.

However, an important goal of AI is for computers to think like humans. AI in Radiology should ultimately be able to integrate and analyze multiple pieces of information, be able to learn, generalize and abstract from a small number of examples and perform sequential cognitive tasks. In other words, human-level intelligence in Radiology AI resembles a control task.  

\subsection*{Early promise, and limitations, of Deep Reinforcement Learning.}

Deep Reinforcement Learning (DRL) is a type of AI that dominates the fields of complex control tasks, namely robotics. This is because it provides the appropriate sampling and implicit learning from the environment in the form of rewards. As such, DRL may help to lead us past the above-described impasse in Radiology AI. In fact, early applications of DRL have shown remarkable ability to generalize based on small training sets \cite{stember2020deep,stember2020reinforcement,stember2020unsupervised,stember2021deep_3d,stember2021deep}. 

However, the Deep $Q$ Networks (DQNs) in DRL are unable to truly converge monotonically to global optima, and off-policy exploration adds randomness; the end result is that DRL can approximate, but not fully achieve, truly optimal policies. This limits clinical deployment in the sense that best policies must be selected out for clinical use. The best policy selection constitutes a type of hyperparameter tuning, the hyperparameter being training time.

For clinical deployment, a network would preferably be able to train and adapt to new "environments" (e.g. MRI settings) and tasks in a continuous, fluid, and essentially never-ending manner. As such, it is not feasible to require an optimized training time, as is typically used in Supervised Deep Learning to avoid overfitting the training set and afflicts DRL because it does not monotonically converge to optimal policies. Requiring a fixed training time would be unfeasible for seamless continuous learning since in the latter there is no clear starting time. 

\subsection*{Evolutionary strategies: history, background and major strengths.}

Evolutionary strategies, inspired by biological evolution, hold the promise to deliver the missing element of monotonic convergence. Pioneered in the 1960’s by Rechenberg and Schwefel \cite{emmerich2018evolution,beyer2002evolution}, evolutionary strategies found early use primarily in engineering, often discovering effective designs that had never before been considered by engineers \cite{emmerich2018evolution}. Evolutionary strategies are known to explore optimization spaces thoroughly and inevitably arrive at global optima \cite{bertsekas2009convex}. 

Since CNNs have been the focus of intense research interest over the past decade, researchers have applied evolutionary strategies to CNNs. This constitutes the field of Deep Neuroevolution (DNE). We seek in the present work to apply Deep DNE to the task of MRI brain sequence classification as a proof-of-concept. 
 
\subsection*{Radiology AI currently depends on Stochastic Gradient Descent.}
 
Essentially all research in Radiology AI, including DRL, relies upon optimizing the CNN parameters, or weights. Researchers do so via stochastic gradient descent (SGD). SGD uses partial derivatives and the Chain Rule of Calculus to systematically adjust the CNN weights toward a set of values that optimizes an objective function, usually a loss between CNN output and a target. 

The problem with SGD is that it assumes a convex, non-noisy feature space, which is essentially never the case; i.e. local optima exist, and can serve as "traps." As such, SGD is not guaranteed to reach the global optimum; the weights often get stuck proceeding toward a local, but not global, optimum. Further, SGD is designed to approach, but not actually reach, these local optima.

\subsection*{Stochastic gradient descent predisposes to data bias and overfitting for small training sets. Transfer learning helps but shows limitations.}

Additionally, SGD-based methods suffer from data bias. The training data that researchers use to tune weights inevitably bias the resulting network. This is because, in order to truly know the optimal surface, an infinite amount of training data would in theory be needed. For the finite, and often small, data sets of Radiology this leads to the notorious problem of severe data bias, commonly referred to as overfitting. 

A typical approach to avoid overfitting small data sets is transfer learning. However, transfer learning is not a panacea. Let us imagine that the source task on which the CNN is pre-trained is not sufficiently similar to the target task to which it will be transfer learned. Or perhaps the relationship is not well leveraged by the transfer method. The result of either scenario can be negative transfer learning. Negative transfer learning is defined as the CNN's performing \textit{worse} after transfer learning \cite{torrey2010transfer,rosenstein2005transfer}. 

There is no clear metric for deciding when the source and target tasks are appropriately matched. Additionally, the need for task matching inherently limits the ability for meta-learning, which is the application of learning on a source task to a significantly different target task and/or environment. If AI is to approach human-level abstraction, understanding, and predictive abilities, it must address all of these challenges \cite{torrey2010transfer}.

As stated perhaps most generally, SGD-based optimization is hindered with regard to continually learning clinical deployment due to the need for hyperparameter tuning. In general, for clinical feasibility, it is preferable not to require a validation set for hyperparameter tuning. Again, this also applies to DRL, given its inability to converge strictly to optimal policies. In that case, training time is the hyperparameter that requires tuning. Additional examples of hyperparameters on which CNN performs depends sensitively are learning rate and momentum. 

Generally speaking, CNN gradients depend on where one is on the optimization surface. Imagine suddenly switching to a new optimization surface (e.g. applying a pre-trained network to transfer learn on images from scanners at a new institution, which may have slightly different machine settings.) The gradients from the pre-trained network now are probably markedly mismatched for the shape of, and position on, the new optimization surface. 

\subsection*{Most modern Radiology AI uses very large (deep) neural networks, but small networks are preferable for clinical deployment.}

In general, small CNNs and small training sets are desirable. Small CNNs have shorter inference times than large CNNs. The difference becomes increasingly pronounced at scale, such that clinical application, which would presumably require millions of deployments, would be profoundly slowed for larger CNNs. The preference for small training sets is intuitive. It is always easier to obtain, process and label small training sets. Furthermore, some image types are inherently sparse, for example, images from patients with rare diseases.

\subsection*{Prior applications of Deep Neuroevolution to Radiology.}

Given its aforementioned strengths, DNE holds the promise to address the challenges of overfitting and the requirement of large CNNs.  

There are two major pathways for DNE in Radiology Deep Learning:
\begin{itemize}
    \item 1) Optimize the CNN architecture/hyperparameter search, then train with Supervised Deep Learning.
    \item 2) Directly optimize the CNN parameters, eschewing Supervised Deep Learning and more generally stochastic gradient descent. Whereas the latter is an approximation for parameter optimization, DNE provides \textit{exact} solutions.
\end{itemize}

Recent work applying DNE to Radiology has focused on the first above strategy. Ahmadian et al. applied DNE to COVID-19 chest x-ray classification \cite{ahmadian2021novel}. Hassanzadeh et al. used DNE for segmentation on CT and MRI \cite{hassanzadeh2021evolutionary}. The researchers in both projects used DNE to tune the CNN \textit{hyperparameters}, but not the \textit{parameters}. 

\section*{Our application of Deep Neuroevolution to MRI sequence classification, and prior approaches to this task.}

In this work, we seek to show the proof-of-principle application of DNE to perform the second strategy from the above: optimally tune CNN \textit{parameters}. This constitutes an alternative to the SGD method of CNN weight adjustment. 

As the first application of our approach, we chose the prediction task of classification on MRI brain scans, specifically multi-class sequence detection. Image classification may be the fundamental task of artificial intelligence (AI) in radiology \cite{mcbee2018deep,saba2019present,mazurowski2019deep,weikert2020practical}. MRI sequence identification is often helpful when attempting to acquire images of a particular sequence when natural language processing fails due to variable naming schemes. Prior supervised deep learning approaches have shown success in this task \cite{noguchi2018artificial,ranjbar2020deep}, reaching up to $99-100 \%$ testing set accuracy. However, they both relied on very large CNNs (ImageNet, GoogleNet, and VGGnet) that had been pre-trained on enormous image databases (ImageNet, 1.2 million images.) Additionally, even the training sets of the target task were somewhat large, both consisting of over 1000 images.

\section*{Methods}

\subsection*{Data collection}

We selected the first 40 training set images from the BraTS 2020 Challenge brain tumor database \cite{menze2014multimodal,bakas2017advancing,bakas2018identifying}. This database consists of primary glioma-containing MRI scans, pre-anonymized and freely downloadable. As we used publicly available images, IRB approval was deemed unnecessary for this study. 

\begin{figure}[H]
\hspace*{-0.2cm}  
\includegraphics[width=12cm]{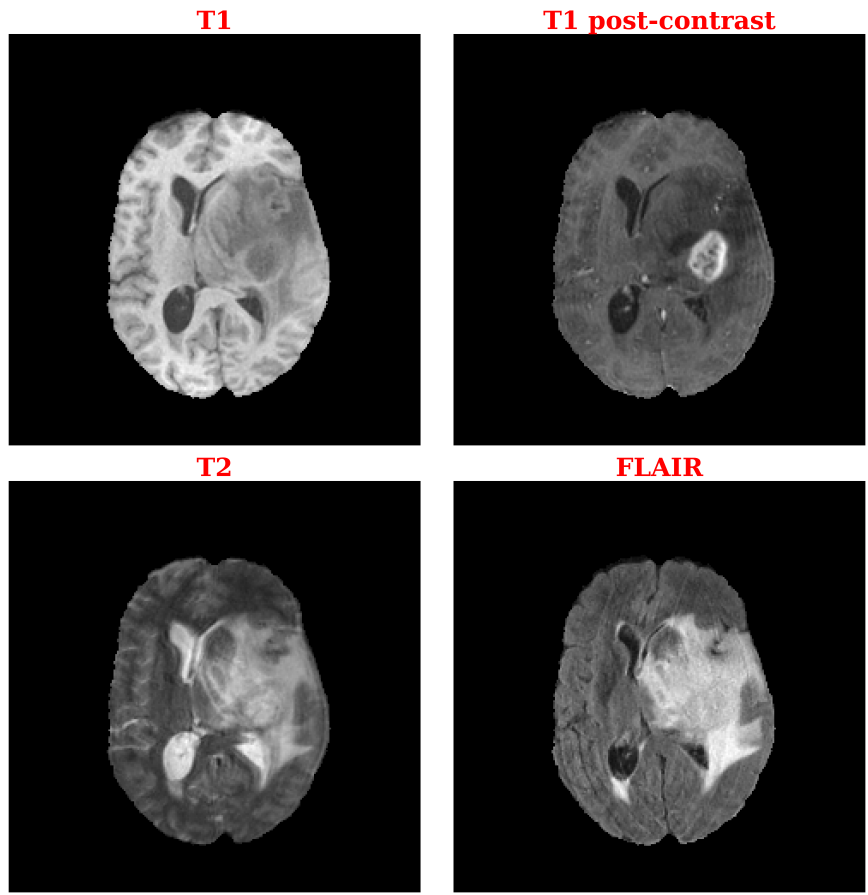}
\caption{Example patient images, four sequences, all mid-slice.}
\label{fig:sequences}
\end{figure}

For each patient, we selected the mid-axial slice for each of the sequences: T1 pre-contrast, T1 post-contrast, T2, and FLAIR. An example for one patient is shown in Figure \ref{fig:sequences}. Many of the mid-slices contained tumors, but some did not (lesion is located superior or inferior to the mid-axial slice for that patient). Since there was no significant difference between the training, validation, and testing sets in the BraTS database, and we only need 40 cases, we took these all from the training set, dividing them even into training and testing sets for our analysis. Noting that each study contains four sequences of interest (listed above), this produced 80 training set images and 80 testing set images.


\subsection*{Convolutional Neural Network (CNN)}
 
\begin{figure}[H]
\hspace*{-0.2cm}  
\includegraphics[width=12cm]{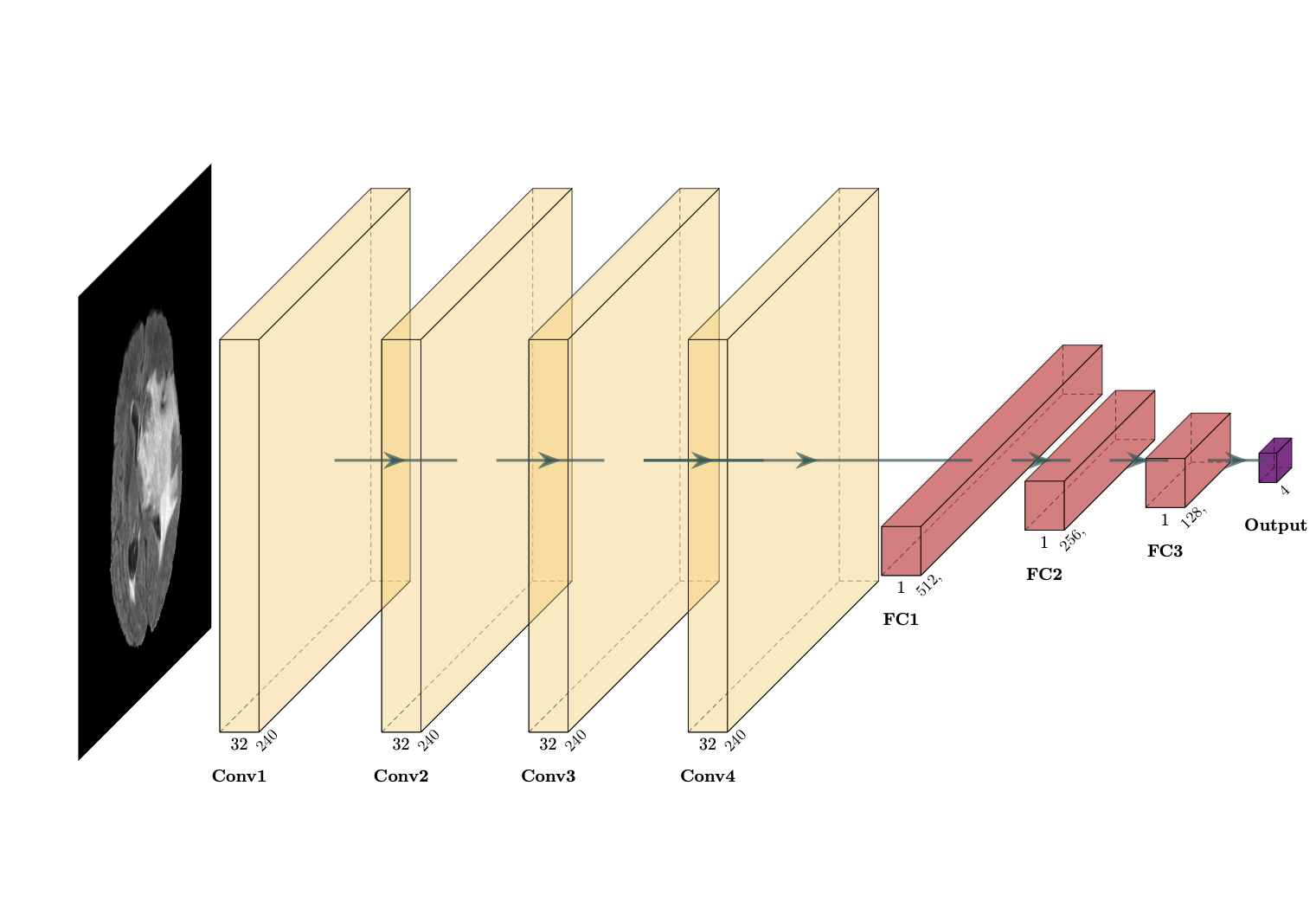}
\caption{Convolutional neural network (CNN) architecture.}
\label{fig:CNN_architecture}
\end{figure}

Like all evolutionary stories, ours begins with a first generation, a CNN whose weights are randomly initialized via the standard Glorot scheme. The CNN consists of four convolutional layers, each with 32 output channels/feature maps. We use kernels/filters composed of $3 \times 3$ weights. We employ a stride of $2$ in the $x$ and $y$ directions, with padding of one applied to the input at each step. ReLu activation follows each 2D convolution. We then flatten the last convolutional layer's output. The resulting flattened vector proceeds to fully connected layers of sizes 512, 256, and 128. We connect this last 128-node layer to the four output nodes representing our four MRI sequences. A schematic of the CNN architecture appears in Figure \ref{fig:CNN_architecture}. 

Of note, we did not attempt to normalize weights or gradients via Dropout, Batch Normalization, or the Softmax function. From our observations, there was no need to do so. We speculate that avoiding SGD averts the commonly encountered problems of exploding or vanishing gradients. 

We also note that the total number of weights in our network is approximately $1.4 \times 10^5$. While this may sound like a large number in general, it constitutes a \textit{small} CNN by the standards of Deep Learning. We make this point more explicitly via the comparisons of Table \ref{table:1}.

\subsection*{Classification accuracy provides the fitness criterion.}

For each training set image, a forward pass of the CNN is performed, and the maximum output node forms the class prediction. If the image is classified correctly, a reward of one is received, otherwise, the reward is zero. We sum the reward for each training set image into a cumulative/total reward. Hence, for the 80 training set images (20 patients, 4 sequences), the maximum and minimum possible total reward for a given CNN is 80 and 0, respectively. This reward, a direct measure of training set sequence classification accuracy, is used as the fitness evaluation criterion. As such, it guides the selection of the best mutations across generations, as expounded upon below and further in the Appendix. 
 
\subsection*{Deep Neuroevolution selects for the fittest mutations and passes them on to future generations.}

Biological evolution, and thus evolutional strategies such as DNE, contain two essential components:
\begin{itemize}
    \item 1) Mutation. 
    \item 2) Selection.
\end{itemize}

These represent the two opposing forces that drive evolution. Random mutations allow a thorough searching through the space of possibilities, here the set of possible CNN weights. Selection, based on a fitness criterion or criteria, gives direction to the parameter sampling that mutation produces. Selection drives the randomly produced parameters from mutation toward meaningful values given the task of interest. 

We may also take the biological perspective; in biological evolution, the changes from mutations are typically harmful, but occasionally adaptive. If they are adaptive for the selective pressures they can propagate in future generations. Analogously, we "mutate" the offspring by random changes in the weights of their CNNs, but keep only those mutations that are helpful toward increasing classification accuracy. 

\begin{wrapfigure}{4}{0.5\textwidth}
  \vspace{-20pt}
  \begin{center}
    \includegraphics[width=0.48\textwidth]{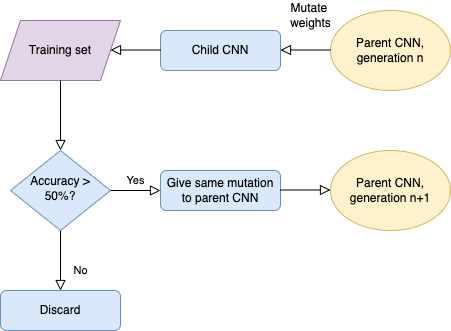}
  \end{center}
  \vspace{-5pt}
  \caption{Flow chart illustrating the update process of CNNs based on mutations and incorporation into the next generation's parent CNN of those mutations that lead to good performance.}
  \label{fig:flow_chart}
  \vspace{-10pt}
\end{wrapfigure}

In our study, mutations will be conferred to "children" CNNs by "parent" CNNs. We will evaluate the fitness of children CNNs via their total training set rewards, which reflect how well the CNNs classify sequences. We will then incorporate the mutations of the better-performing child CNNs into the "genome" of the parent CNN. The mutation that has the most influence in this incorporation will be that of the best-performing child CNN, followed by the mutation of the second best-performing child, etc. This newly-mutated parent will become the parent of the next generation. We choose 40 children per generation (after the first generation, which has only one parent CNN). The process will repeat for many generations, in our cases over 50,000, until the system reaches global convergence.

If a particular mutation is advantageous, the fitness will be evaluated by the total reward that the CNN obtains, as stated above. The children's CNN mutations are noted. The parent essentially undergoes a combination of the best mutations, as observed in the set of offspring. Then the mutated/updated parent becomes the parent of the next generation, producing a new set of 40 randomly mutated child agents, and the process continues for thousands of generations. 

Herein, a source of possible confusion arises, a break from the analogy with biology that may warrant clarification: The children themselves do not propagate in the next generation. However, they do propagate in the sense that we incorporate their best mutations into the next generation's parent CNN. The process, illustrated for one child, can be seen in Figure \ref{fig:flow_chart}.

More technical formulations of the procedure are provided in the Appendix. 

\subsection*{Deep Reinforcement Learning (DRL) classification for comparison}

Because recent research on DRL showed high accuracy for small training sets, we seek to compare the performance of DRL with DNE. To do so, we train a DRL-based classifier on the same $80$ training set images as for DNE. We do so via a Deep $Q$ Network (DQN) with architecture essentially identical to that of the CNN evaluated by DNE. We train the DQN weights via SGD. To sample from the image environment, we perform DQN training in tandem with temporal difference learning in an essentially identical manner to that of recent DRL-based binary classification work \cite{stember2021deep,stember2021deep_3d}. 

\section*{Results}

\begin{figure}[H]
\hspace*{-1.5cm}  
\includegraphics[width=15cm]{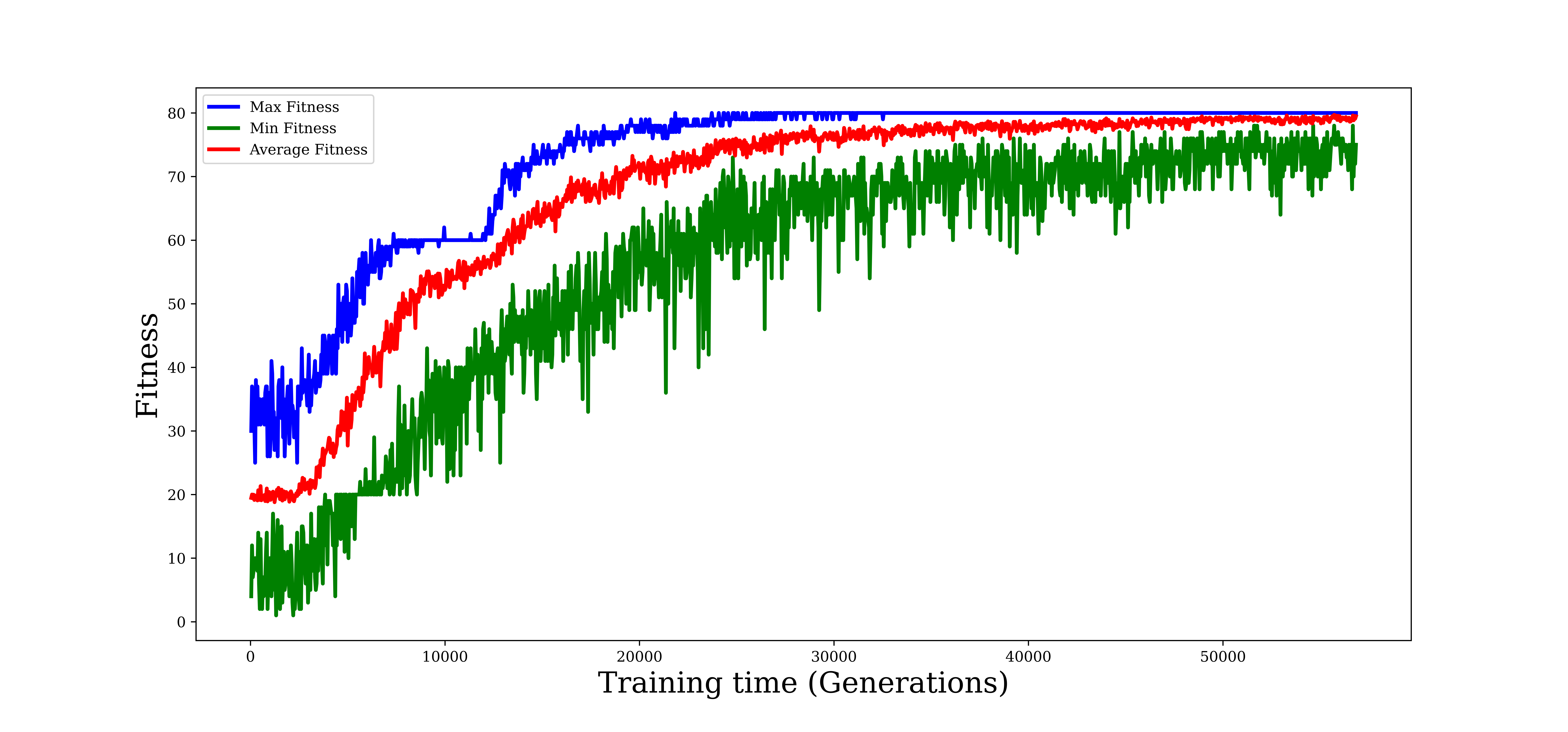}
\caption{Training set accuracy as a function of time in units of generations. The accuracies are for the children CNNs. The best-performing child (blue) is plotted along with the average of children (red), as well as the worst-performing child (green). These are ultimately all seen to converge on perfect accuracy (fitness of 80/80 correctly predicted image sequences.)}
\label{fig:training}
\end{figure}

The training time of DNE consisted of over 50,000 generations of evolution, after which global convergence was clearly manifest. The training time was 20 hours and 14 minutes on an 8-core CPU (no GPU used) in Google Colab Pro. Because we did not perform backpropagation of the CNN (no SGD), we did not require a GPU. We also did not perform any particular hyperparameter tuning at any point during training.

\begin{figure}[H]
\hspace*{-1.5cm}  
\centering
\includegraphics[width=15cm,height=15cm]{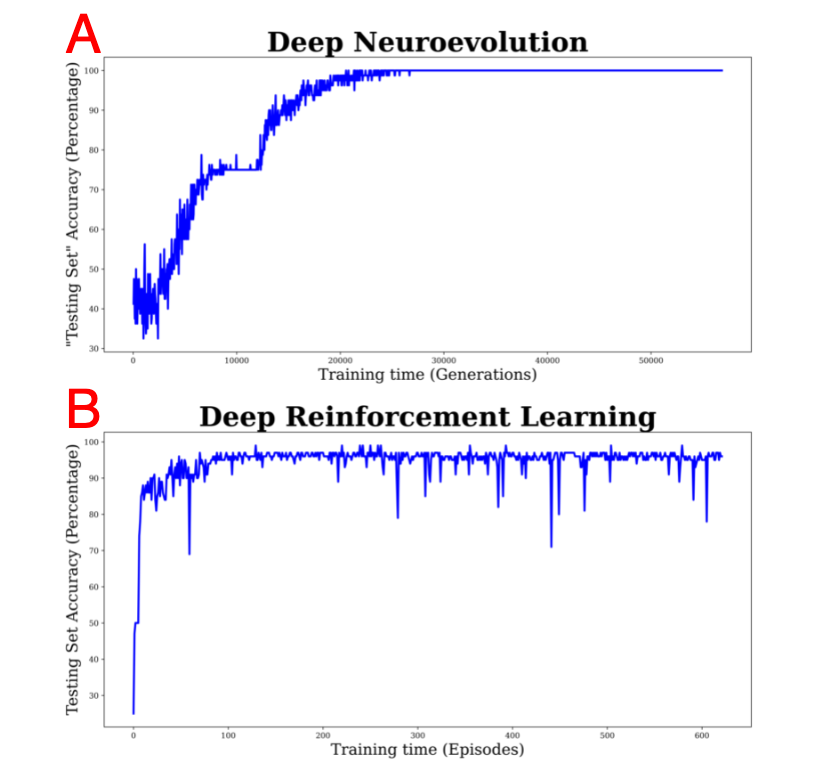}
\caption{ Testing set accuracy as a function of time in units of generations. Accuracy is evaluated as the maximum value over the 40 children CNNs and single parent CNN.}
\label{fig:testing}
\end{figure}

Training results for the child CNNs are shown in Figure \ref{fig:training}. We see that DNE converges strictly and monotonically to the global optimum, achieving 100$\%$ accuracy. Not only the top-performing child agent achieves this result, but eventually, the average performance and even the lowest-performer converge on the optimum. This follows from the fact that the parent CNN incorporates the children's mutations in a way that emphasizes the best performers. This "drags along" the lower-performing children because in later generations they are beginning from a more adapted starting point. 

The application of the parent and child CNNs to the separate testing set is shown in Figure \ref{fig:testing}A, which plots the maximum accuracy versus training time. As a comparison to DNE, DRL's performance on the same training/testing data is shown in Figure \ref{fig:testing}B. DRL approaches but does not fully converge on the optimal policy. Instead, it oscillates around testing set accuracy values in the upper 90 percent range. 

Table \ref{table:1} provides a comparison between DNE, DRL, and prior Supervised Deep Learning using transfer learning approaches to sequence identification \cite{noguchi2018artificial,ranjbar2020deep,khan2020survey}. Although we have the smallest training set and CNN, DNE achieves 100$\%$ testing set accuracy. 

\section*{Discussion}

\subsection*{Reiterated overview of the various algorithms/approaches.}

Revisiting the conversation broached in the Introduction section, we may consider the various types of AI and optimization as follows:
\begin{itemize}
    \item SGD in SDL represents monotonic guidance with an explicit objective function (the target loss). For this to work well, one needs huge amounts of data to define adequately the objective function surface. Additionally, problems such as exploding/vanishing gradients are present, which could be exacerbated by sub-optimal tuning of the learning rate. Training is likely to get stuck in a local optimum, and never even in fact reach \textit{that}.
    \item SGD in DRL represents monotonic guidance with an implicit/stochastic objective function. The stochasticity comes from off-policy contributions to the replay memory buffer, which we randomly sample for DQN training. In other words, SGD for DRL is chasing a moving target, the temporal difference loss. 
    \item DNE represents gradient-free optimization, stochastic guidance with explicit objective function (fitness). It can take arbitrarily sized steps in arbitrary directions in its exploration of the objective function surface. It is guaranteed to converge to the global optimal, \textit{eventually}.
\end{itemize}

\begin{table}[H]
\begin{tabular}{ |p{2cm}|p{1.3cm}|p{1.0cm}|p{1.7 cm}|p{1.6cm}|p{1.2cm}|  }
\hline
\multicolumn{6}{|c|}{Comparison of AI Sequence Detection Methods} \\
\hline
Paper & Accuracy ($\%$) & Pre-trained & Model & $\#$ Weights ($\times 10^6)$ & Training set size \\ 
\hline
Noguchi et al & 74 & yes & AlexNet & 61 & 384 \\
Noguchi et al & 100 & yes & GoogLeNet & 7 & 384 \\
Ranjbar et al & 99 & yes & VGGNet & 138 & 9,600 \\
\textbf{ours} & 99 & no & DRL & 0.14 & 80 \\
\textbf{ours} & 100 & no & DNE & 0.14 & 80 \\

\hline
\end{tabular}
\caption{Comparison of AI sequence detection methods. Abbreviations: DRL = Deep Reinforcement Learning, DNE = Deep Neuroevolution. DRL and DNE reach high and perfect testing set accuracy, respectively, without pre-training, and using much smaller networks as well as fewer training images. This is not fully expressed in these numbers because ImageNet, used for pre-training, contains approximately 1.2 million images.  }
\label{table:1}
\end{table}

\subsection*{Advantages of Deep Neuroevolution.}

DNE has the potential to provide the most reliable possible AI algorithms, train on small training sets, and be easily deployed because it works with small CNNs. We see that DNE reaches perfect training and testing set accuracies with small training sets. No hyperparameter tuning is necessary for this result. No tricks to reduce overfitting, such as Dropout, Batch Normalization, or Softmax function on the output were needed.

\subsection*{Drawbacks, limitations, and future directions.}

One clear drawback with DNE is the sheer time required for training, in this relatively simple task exceeding 24 hours. The promising upside is that the update process outlined in Figure \ref{fig:flow_chart} is highly parallelizable. We employed parallel processing with the threads available in the single CPU from Google Colab. However, in future work, with access to large numbers of CPUs, we anticipate significant improvements in training time. 

Although we have emphasized the advantages of the mutation/selection-based parameter search of DNE, SGD does have some important advantages that can, in future iterations, help to increase the speed of DNE training. SGD provides directionality, sorely needed in DNE training. We anticipate that future iterations of DNE will incorporate SGD-derived directions to guide parameter search more efficiently.  

We also anticipate the integration of DNE with DRL. By DRL selection of intelligent actions correlated sequentially, and DNE optimization of DQN-predicted actions, we hope to master complex radiological control tasks, bringing us closer to true human-type intelligence in radiology.

\section*{Conflicts of interest}

The authors have pursued a provisional patent based largely on the work described here.

\section*{Funding}

We gratefully acknowledge external support from the Radiological Society of North America (RSNA) and the American Society of Neuroradiology (ASNR).

\section*{Appendix}

\appendix

\section{Perturbing the child agents}

Let us denote the parent CNN weights by $\vec{w} = \{ w_i \}_{i=1}^N$. In the case of our CNN, shown in Figure \ref{fig:CNN_architecture}, the number of weights, or cardinality of $\vec{w}$, is $\vert \vec{w} \vert = N_{\text{w}} = 1.4 \times 10^5$. 

In general, to produce $N_{\text{offspring}}$ child CNNs, we mutate or perturb $N_{\text{offspring}}/2$ of the following two transformations:
    \begin{align}
    \label{perturb1}
        \vec{w}_p^{(+)} =& \  \vec{w} + \vec{\Delta}^{(+)} = \  \vec{w}+\sigma \times \vec{\epsilon}  \\
        \label{perturb2}
        \vec{w}_p^{(-)} =& \  \vec{w} + \vec{\Delta}^{(-)} = \  \vec{w}-\sigma \times  \vec{\epsilon},
    \end{align}
where $\Delta$ is the overall perturbation (with superscript indicating direction), $\sigma$ is a weighting factor and $\vec{\epsilon}$ is a random perturbation vector. The elements of $\vec{\epsilon}$ are selecting according to a normal random distribution. 

In our case, in each generation the parent produces $N_{\text{offspring}} = 40$ children. Hence, performing the perturbation from Equations \ref{perturb1} and \ref{perturb2}, we have the $40 \times N_{\text{w}}$ matrix $\mathcal{W}_p$. 

\section{Updating the parent agent}

We update the unperturbed parent CNN weight vector at generation $n$ according to the following weighted sum of child CNN weight perturbations:
\begin{equation}
\label{update_parent}
    \vec{w}_{n+1} = \vec{w}_n + \frac{\text{lr}}{\sigma N_\text{c} } \sum_{c=1}^{N_c} \text{sign}(\vec{\Delta}) R_c^S\vec{\epsilon},
\end{equation}
where $\text{lr}$ is the learning rate and $R_c^S$ is the "shaped" return. We used a learning rate value of 0.1, without decay. Although the learning rate serves here as a scaling of the weight updates, we can not overstate the difference between the learning rate in DNE and the learning rate from SGD. DNE is sure to converge to a global optimum regardless of the learning rate. In contrast, SGD's local convergence prospects are highly sensitive to the learning rate value chosen. 

Let us denote by $\vec{R}$ the set of training set rewards for the child agents, of length $\vert \vec{R} \vert = N_{\text{c}}$, recast in descending order. Then $\text{ind}_{R_c}$ is the index of the first occurrence of the reward value of child agent $c$ in $\vec{R}$. Finally, the shaped return is given by
\begin{equation}
\label{shaped_return}
   R_c^S = \frac{ max \{ 0, \log_{2} \left( \frac{N_c}{2}-1 \right) - \log_{2} \left( \text{ind}_{R_c}-1 \right) \} } {\sum_{c=1}^{N_c} max \{ 0, \log_{2} \left( \frac{N_c}{2}-1 \right) - \log_{2} \left( \text{ind}_{R_c}-1 \right) \} }.
\end{equation}
The denominator is a normalizing factor, similar to that in the Softmax function. The intuition behind the numerator of Equation \ref{shaped_return} is that the rewards in the lower half of the range of possible values are excluded from contributing to the update of the parent weights in Equation \ref{update_parent}. This is because, for these lower-reward CNNs, $R_c^S=0$. Hence only the child CNNs that perform well have their perturbation added to the parent weights, and the larger the reward, the higher this contribution. By analogy with biology, we can think of this as the parent observing offspring performance, and incorporating their mutations into the parent's own genome, but with the best mutations getting expressed the most. Hence, there is a combination of variability, getting contributions from a range of well-performing but not best-performing children, but giving more weight to those that perform the best by the selection metric. Also, the parent can assume that poorly performing offspring have maladaptive mutations, and thus will not incorporate those mutations into the parent's own genome. 

Another consequence of Equation \ref{update_parent} and \ref{shaped_return} is that the parent stops updating its weights when at the global optimum. At this point, the plus and minus signs of the perturbations will cancel each other out in all directions, because their rewards will be of equal magnitude. 

The most compute-intensive process of our evolutionary scheme is the forward passing of each image from the training and testing sets through the CNNs. Because these forward passes are independent for the various child CNNs, this is amenable to parallel processing. As such, we assigned each CNN forward pass through training and testing sets to occur via a different processing thread. As each thread in the CPU became free, a new forward-pass process was assigned to the thread as soon as its prior process was complete.






\printbibliography

\end{document}